\documentclass[lettersize,journal]{IEEEtran}
\usepackage{amsmath,amsfonts}
\usepackage{algorithmic}
\usepackage{amsthm}
\usepackage{pifont}
\usepackage{algorithm}  
\usepackage{hyphenat} 
\usepackage{array}
\usepackage{booktabs} 
\usepackage[caption=false,font=normalsize,labelfont=sf,textfont=sf]{subfig}
\usepackage{textcomp}
\usepackage{stfloats}
\usepackage{url}	
\usepackage{verbatim}
\usepackage{graphicx}
\usepackage{cite}
\newtheorem{theorem}{Theorem}[section]
\newtheorem{lemma}{Lemma}[section]
\theoremstyle{plain}

\begin{document}

\title{AlignFed: Alignment-Aware Asynchronous Federated Fine-Tuning for Large Language Models in Heterogeneous Edge Environments}

\author{Yan Wang, Ziyi Gao, Rui Wang*
     
	\thanks{Yan Wang, Ziyi Gao and Rui Wang are with the Department of Computer and Communication Engineering, University of Science and Technology Beijing, Beijing 100083, China(e-mail: D202210405@xs.ustb.edu.cn; m202410658@xs.ustb.edu.cn; wangrui@ustb.edu.cn).}
}

\maketitle

\begin{abstract}
Large Language Models (LLMs) have significantly propelled the advancement of edge intelligence and have been widely deployed across various scenarios, including autonomous driving, industrial inspection, and personalized IoT services. However, the collaborative adaptation of LLMs on edge devices continues to face formidable challenges due to strict data privacy constraints, highly heterogeneous computing and communication resources, and the non-independent and identically distributed (non-IID) nature of local data. Federated Fine-Tuning (FFT) enables the collaborative optimization of distributed models without exposing raw data. Yet, traditional synchronous aggregation suffers from a severe "straggler effect," resulting in high system latency and low resource utilization. Existing asynchronous federated learning methods are predominantly designed for small-to-medium-scale models and struggle to address the specific challenges inherent in LLM fine-tuning—namely, model drift caused by stale updates, aggravated client drift stemming from data heterogeneity, and aggregation fairness imbalance resulting from the dominance of fast clients. To address these issues, this paper proposes AlignFed, an asynchronous federated fine-tuning framework for LLMs tailored to heterogeneous edge environments. AlignFed employs a lightweight multi-stage semantic alignment mechanism comprising three core modules: version-aware update grouping, cross-version semantic alignment based on a mini-batch calibration set, and fairness-aware aggregation that integrates both update freshness and client participation frequency. This framework effectively mitigates cross-version model drift and client drift while enhancing aggregation fairness, thereby achieving stable and efficient asynchronous federated optimization in scenarios characterized by high heterogeneity and significant update staleness. Experimental results across multiple benchmark datasets and two mainstream LLM backbones demonstrate that, compared to existing synchronous and asynchronous baseline methods, AlignFed achieves faster and more stable convergence, superior robustness against stale updates, and enhanced generalization performance.
\end{abstract}


\section{Introduction}
\IEEEPARstart{T}{he} rapid development of edge intelligence is transforming numerous mobile and IoT applications, including autonomous driving, industrial inspection, and personalized IoT assistants~\cite{sathyam2025foundation, zhang2024edgeshard, shen2024large}. Recent advances in large language models (LLMs) further enhance the capabilities of these applications by enabling complex reasoning and multimodal interactions. However, deploying and adapting LLMs on edge devices remains extremely challenging due to three key limitations: strict data privacy (prohibiting the upload of raw data), resource heterogeneity (clients have varying computational and communication capabilities), and skewed data distributions (non-independent and identically distributed local datasets)\cite{10852519,9496155}.

Federated learning (FL) offers a promising paradigm by enabling collaborative model training without sharing raw data~\cite{sun2024improving, 10998940}. Building upon FL, federated fine-tuning (FFT) enables large foundation models to be adapted across distributed edge clients. However, most existing FFT methods rely on \emph{synchronous aggregation}, where the server must wait for all selected clients to finish local updates before proceeding. In heterogeneous edge environments, this leads to severe \emph{straggler effects}~\cite{zhang2025fedraa}, resulting in high latency and poor resource utilization~\cite{ji2021computation,11352945}.

In real-world edge environments, the computational and communication capabilities of devices vary significantly. For example, lightweight IoT boards (such as Raspberry Pi and Jetson Nano) operate several orders of magnitude slower than industrial-grade edge servers or GPU-equipped devices. This imbalance leads to the so-called “straggler effect,”~\cite{ji2021computation} where slower devices delay the entire aggregation process. As a result, synchronous training suffers from increased system latency, low resource utilization, and delayed global model updates~\cite{samarakoon2019distributed}. This motivates the adoption of asynchronous aggregation to improve timeliness and system efficiency~\cite{xu2023fedlc, xie2019fedasync, nguyen2022fedbuff}.

However, existing AsyncFL methods~\cite{gul2025syncfed,forootani2025asynchronous} are primarily developed for traditional CNN/MLP models with dense full-parameter updates, and are not well-suited for LLM fine-tuning under low-rank adaptation. 
This limitation stems from two fundamental structural mismatches: 
(1) Traditional DNNs rely on full-parameter gradient updates, whereas LLM fine-tuning typically adopts low-rank adaptations (e.g., LoRA~\cite{hu2022lora}), where updates lie in a constrained low-dimensional subspace. Direct asynchronous aggregation may disrupt this low-rank structure and lead to degraded adaptation quality. 
(2) The representation space of LLMs is highly structured and semantically sensitive~\cite{jiang2024origins}, making it vulnerable to version inconsistency. In asynchronous settings, stale updates computed from different model versions can introduce misaligned optimization directions, resulting in unstable convergence and performance degradation.

These structural mismatches manifest in several fundamental challenges when extending AsyncFL to LLM fine-tuning in heterogeneous edge environments:

First, asynchronous training inevitably generates stale updates from clients trained on different model versions (i.e., staleness accumulation). This issue is further amplified in LLM scenarios: due to the high dimensionality and complex parameter space of LLMs, direct aggregation of these stale updates is more likely to trigger training instability and significant model drift~\cite{zhang2025orthogonal}.

Second, asynchronous training exacerbates heterogeneity amplification. In LLM fine-tuning, edge devices typically hold highly non-independent and identically distributed \cite{9496155} (non-IID) data; coupled with the more pronounced differences in the magnitude of LLM parameter updates, local updates tend to converge toward different local optima. This not only induces severe client drift but also leads to inconsistent update directions (directional bias and drift) across clients, ultimately degrading model performance~\cite{solans2024non, chang2025mitigating}.

Third, imbalanced client participation can bias the global model. Faster clients contribute updates far more frequently than slower ones, resulting in the global model being dominated by this small subset of fast clients. To address this, the issue of unfair participation under heterogeneous devices needs to be explicitly resolved.

To systematically mitigate the above challenges—model drift induced by staleness, client drift caused by heterogeneity amplification, and unfair aggregation due to imbalanced participation—this paper proposes AlignFed, a novel asynchronous federated fine-tuning framework for large language models. AlignFed introduces a multi-stage alignment mechanism consisting of three key components:

(i) version-aware update grouping, which explicitly organizes updates according to model versions to manage stale updates;

(ii) cross-version semantic alignment, which maps stale updates into a consistent representation space before aggregation using a lightweight calibration set; and

(iii) fairness-aware aggregation, which adaptively weights updates based on update freshness, update strength, and participation frequency to balance client contributions.

By jointly addressing model drift, client drift, and unfair aggregation, AlignFed enables stable and fair convergence for asynchronous federated LLM fine-tuning in heterogeneous edge environments.

\noindent\textbf{Our main contributions are summarized as follows:}
\begin{itemize}

\item We propose AlignFed, the first asynchronous federated fine-tuning framework specifically tailored for LLMs on heterogeneous edge devices.

\item We design a multi-stage alignment mechanism that integrates version-aware update grouping, cross-version semantic alignment, and fairness-aware aggregation, effectively mitigating issues such as model drift, client drift, and unfair aggregation in asynchronous training.

\item We provide theoretical guarantees showing that AlignFed achieves bounded-drift convergence under asynchronous aggregation.
\end{itemize}

\section{Related Work and Preliminaries}
This section reviews the background and related research relevant to asynchronous federated fine-tuning of LLMs in heterogeneous edge environments. We first discuss asynchronous federated learning, followed by parameter-efficient federated fine-tuning approaches, and finally review studies on semantic consistency and fairness in federated aggregation.
\subsection{Asynchronous Federated Learning}

Existing AsyncFL methods aim to mitigate stragglers by relaxing synchronization constraints. Representative work such as FedAsync~\cite{xie2019fedasync} allows the server to update the global model immediately upon receiving a client update, thereby reducing waiting time caused by slower devices. FedBuff~\cite{nguyen2022fedbuff} further improves aggregation efficiency by buffering a sufficient number of updates before triggering a global update. They improve training efficiency via immediate or buffered aggregation.

Subsequent studies have focused on improving the robustness and convergence behavior of AsyncFL from multiple perspectives. One line of work introduces staleness-aware aggregation strategies, where client updates are reweighted based on their delay to mitigate the negative impact of outdated information~\cite{gul2025syncfed}. Another line incorporates delay compensation mechanisms~\cite{wang2024fadas} or bounded-staleness constraints~\cite{ma2024fedstaleweight} to stabilize asynchronous optimization dynamics.

Beyond aggregation design, recent efforts further investigate the theoretical properties of AsyncFL. For example, convergence guarantees have been established under convex and non-convex settings~\cite{forootani2026asynchronous, forootani2025asynchronous}, while other studies explore discrepancy-aware aggregation to address non-IID data distributions~\cite{liu2025feddm}. These advances collectively improve the stability and applicability of AsyncFL in heterogeneous environments. 

These methods fundamentally assume that client updates are compatible within a shared parameter space, and that aggregation operations are based on the full model update. However, in LLM fine-tuning, updates are parametrically efficient (e.g., LoRA~\cite{hu2022lora} for Low-Rank Adaptation), not full model gradients, and operate in a high-dimensional, tightly coupled parameter space with non-IID data. Therefore, outdated updates are not merely outdated—they are semantically inconsistent. This renders traditional weighted methods that account for outdatedness inadequate.

Specifically, two fundamental mismatches between existing AsyncFL methods and LLM fine-tuning further highlight their incompatibility: First, traditional AsyncFL relies on dense full-parameter updates for CNN/MLP models, whereas LLM fine-tuning adopts low-rank LoRA updates confined to a low-dimensional subspace. Direct application of asynchronous aggregation (e.g., weighted averaging in FedAsync, buffered aggregation in FedBuff) disrupts the low-rank structure of LoRA updates, leading to semantic degradation and poor adaptation quality. Second, existing AsyncFL methods adopt a simple Euclidean space assumption for aggregation, ignoring the highly structured and semantically sensitive nature of LLM representation spaces~\cite{jiang2024origins}. Stale updates from different model versions in asynchronous settings cause misaligned optimization directions, which not only amplifies model drift but also exacerbates client divergence under non-IID data distributions.

These observations indicate that directly applying existing AsyncFL mechanisms to LLM fine-tuning may lead to inconsistencies in representation alignment, amplified client drift, and biased aggregation behavior.

\subsection{Federated Fine-Tuning with Parameter-Efficient Adaptation}

FFT reduces communication and computation overhead by transmitting lightweight parameter updates instead of full-model gradients~\cite{liu2025adaptive}. Parameter-efficient fine-tuning (PEFT) methods such as FedPETuning~\cite{yuan2023fedpetuning} and FedLoRA~\cite{zhang2025fedlora} introduce parameter decomposition techniques to enable personalized federated learning.

By fine-tuning only low-rank components and transmitting these lightweight matrices, such approaches significantly reduce communication costs compared to full-parameter federated training.

Recent studies further improve communication efficiency. FFA-LoRA~\cite{sun2024improving} freezes a subset of adapter parameters to minimize update size. FLASC~\cite{kuo2024federated} introduces sparse adapter communication, while LoRI~\cite{zhang2025lori} employs task-specific sparse masking. FedSA-LoRA~\cite{guo2025selective} separates shared and personalized update components.

Despite these advances, most existing federated PEFT approaches rely on synchronous aggregation. In heterogeneous edge environments, this synchronization requirement forces faster clients to wait for slower ones, reducing system efficiency and increasing end-to-end training latency.

\subsection{Heterogeneity and Fairness in Federated Aggregation}

Client heterogeneity is a fundamental challenge in federated learning, especially when local data distributions are highly non-IID~\cite{solans2024non}. Under such conditions, local updates may drift toward different optimization directions, leading to client drift and degraded global model performance~\cite{chang2025mitigating}. This issue becomes even more pronounced in asynchronous training settings, where stale updates and non-IID data jointly amplify divergence among client updates.

Several studies have explored techniques to mitigate representation inconsistency and model divergence. For example, knowledge distillation~\cite{wu2022communication}, representation matching~\cite{mostafa2019robust}, and feature alignment~\cite{yu2021fed2} have been proposed to maintain consistency among distributed models. However, these methods are primarily designed for synchronous training scenarios and assume homogeneous update timing across clients, making them difficult to directly apply to asynchronous federated fine-tuning of LLMs.

Fairness is another critical concern in federated aggregation, as the global model may become biased toward clients with larger datasets or more frequent participation. Existing studies~\cite{mohri2019agnostic,li2019fair} address this issue by reweighting training objectives or optimizing worst-case client performance. More recent approaches introduce fairness-aware optimization algorithms~\cite{chen2024fair,zeng2024fair} to mitigate domain skew and participation imbalance.

Nevertheless, most fairness-aware federated learning methods assume synchronous participation and full-parameter training. In asynchronous environments, clients with stronger hardware or more stable network conditions naturally contribute updates more frequently, which further amplifies participation imbalance.

\section{AlignFed Framework and Problem Formulation}
\label{sec:method}
In this section, we introduce the proposed \textbf{AlignFed} framework for asynchronous federated LLM fine-tuning with LoRA. We first formalize the global and local optimization objectives and define core notations, then present the overall pipeline of AlignFed, and finally detail its three key components and theoretical analysis.

\subsection{Problem Formulation and Notation}
We consider an asynchronous federated learning system for LLM fine-tuning (with LoRA as the lightweight fine-tuning paradigm for implementation), consisting of a central server and a set of participating clients denoted by $\mathcal{C} = \{1, 2, \dots, N\}$. Each client $i \in \mathcal{C}$ owns a private dataset $\mathcal{D}_i = \{(\boldsymbol{x}_{i,j}, y_{i,j})\}_{j=1}^{|\mathcal{D}_i|}$ for the same downstream task (e.g., code generation, natural language reasoning), where the data in $\mathcal{D}_i$ is drawn from a non-IID and heterogeneous distribution. All clients perform lightweight fine-tuning via LoRA, which freezes the pre-trained LLM backbone $W_{\text{backbone}}$ and only optimizes low-rank adapter parameters $A_i, B_i$ (rank $r \ll d$, $d$ is the LLM hidden dimension), making the local model $W_i = W_{\text{backbone}} + A_i B_i^T$.

\subsubsection{Optimization Objectives}
The core goal of AlignFed is to learn a global LoRA adapter such that the global model minimizes the aggregated empirical risk over all clients' private datasets, with the constraint of asynchronous lightweight fine-tuning.
\paragraph{Global Objective Function}
The global optimization objective is defined as:
\begin{equation}
	\begin{aligned}
	&\min_{W_{\text{global}}^A, W_{\text{global}}^B} \mathcal{F}(W_{\text{global}}^A, W_{\text{global}}^B) = \\
	& \frac{1}{N} \sum_{i=1}^N \mathbb{E}_{(\boldsymbol{x}, y) \sim \mathcal{D}_i} \left[ \ell\left( \text{LLM}(W_{\text{backbone}} + W_{\text{global}}^A (W_{\text{global}}^B)^T; \boldsymbol{x}), y \right) \right]
	\label{eq:global_obj}
	\end{aligned}
\end{equation}
where $\ell(\cdot, \cdot)$ is the task-specific loss function (cross-entropy, negative log-likelihood, etc.), and $\text{LLM}(\cdot; \boldsymbol{x})$ denotes the forward pass of the pre-trained LLM with input $\boldsymbol{x}$.

\paragraph{Local Objective Function}
For each client $i$, we design a local objective with representation consistency regularization to mitigate client drift from the global model, which is critical for asynchronous federated training \cite{forootani2025asynchronous}:
\begin{equation}
	\begin{aligned}
	&\min_{A_i, B_i} \mathcal{F}_i(A_i, B_i) = \\ &\mathbb{E}_{(\boldsymbol{x}, y) \sim \mathcal{D}_i} \left[ \ell\left( \text{LLM}(W_{\text{backbone}} + A_i B_i^T; \boldsymbol{x}), y \right) \right] + \lambda \|\phi_i - \phi_g^{(v_i)}\|_2^2
	\label{eq:local_obj}
	\end{aligned}
\end{equation}
where $\phi_i/\phi_g^{(v_i)}$ are the local/global feature representations extracted from the LLM's penultimate layer, and $\lambda > 0$ is the regularization coefficient balancing task performance and representation consistency.

\subsubsection{Key Notations}
The server maintains a versioned global model to track asynchronous updates. Let $W_g^{(t)} = W_{\text{backbone}} + (W_{\text{global}}^A)^{(t)} (W_{\text{global}}^B)^{(t)T}$ denote the global model at global version $t$ (incremented only when aggregation is performed). Client $i$ starts local fine-tuning from a stale global model snapshot $W_g^{(v_i)}$ ($v_i \le t$) and uploads the LoRA parameter increment $\Delta_i = (A_i - A_g^{(v_i)}, B_i - B_g^{(v_i)})$ with the version tag $v_i$. Table~\ref{tab:notations} summarizes the key notations.

\begin{table}[h]
	\centering
	\caption{Key Notations Used in the AlignFed Framework}
	\label{tab:notations}
	\begin{tabular}{ll}
		\toprule
		\textbf{Symbol} & \textbf{Description} \\
		\midrule
		$\mathcal{C}$ & Set of participating clients $\{1,2,\dots,N\}$ \\
		$\mathcal{D}_i$ & Private local dataset of client $i$ \\
		$W_g^{(t)}$ & Global LLM model (backbone + LoRA) at version $t$ \\
		$v_i$ & Global version index for client $i$'s local fine-tuning \\
		$\Delta_i$ & Client $i$'s lightweight LoRA parameter increment \\
		$\mathcal{G}_k$ & Group of updates originating from global version $k$ \\
		$\phi_i, \phi_g^{(v_i)}$ & Local/global LLM feature representations (penultimate layer) \\
		$\mathcal{D}_c$ & Small server-side calibration set for semantic alignment \\
		$\mathcal{T}_{k \rightarrow t}$ & Linear transformation for cross-version semantic alignment \\
		$\alpha_i$ & Normalized aggregation weight for client $i$’s update \\
		$\omega(v_i)$ & Version-based freshness weight for stale updates \\
		$\xi_i$ & Fairness weight based on client $i$'s participation frequency \\
		$\tau_i = t - v_i$ & Staleness of client $i$'s update (time lag from global version $t$) \\
		\bottomrule
	\end{tabular}
\end{table}

\subsection{Framework Overview}
We propose the \textbf{AlignFed} framework, an asynchronous federated LLM fine-tuning method designed to address three core challenges amplified in asynchronous LLM scenarios: 
\textbf{accumulation of stale updates}, \textbf{heterogeneity amplification}, and \textbf{unfair participation caused by fast clients}. 
Unlike conventional asynchronous federated learning that directly aggregates unaligned local updates, AlignFed introduces a dedicated multi-stage alignment mechanism to sequentially resolve version inconsistency, directional bias, and contribution imbalance.

\begin{figure}[!htbp]
	\centerline{\includegraphics[width=0.4\textwidth]{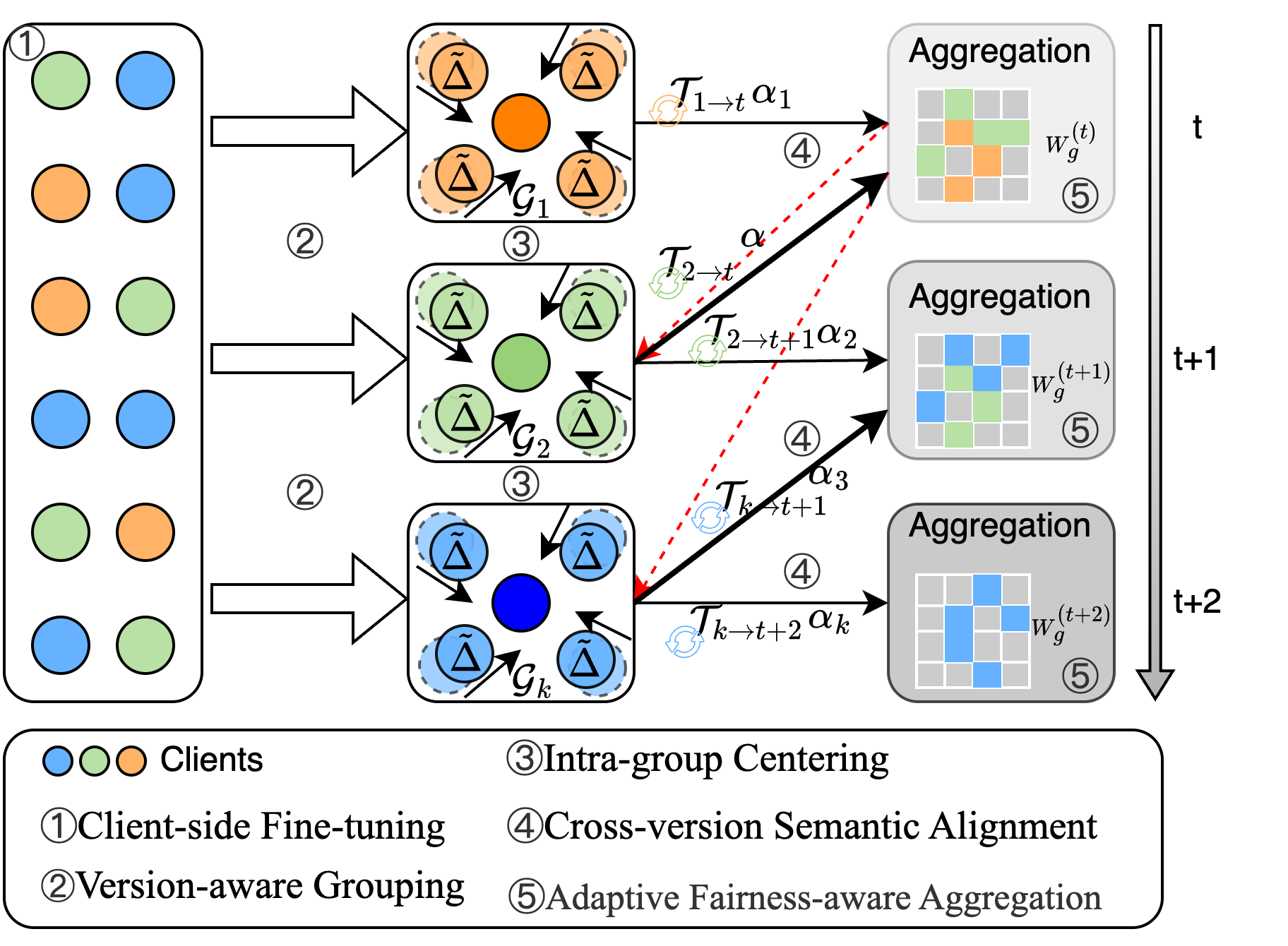}}
	\caption{Overall architecture of the proposed \textbf{AlignFed} framework.}
	\label{framework}
\end{figure}

The overall workflow of AlignFed is illustrated in Fig.~\ref{framework} and summarized in Algorithm~\ref{alg:AlignFed}. The framework contains three core components that correspond one-to-one to the three key challenges:

\begin{itemize}
	
\item \textbf{Version-aware Grouping \& Cross-version Semantic Alignment}: Explicitly handles \textbf{staleness accumulation} (which induces \textbf{model drift}) by grouping updates from different model versions and aligning stale updates to the current semantic space, thus suppressing global model deviation caused by outdated local updates.

	\item \textbf{Intra-group Centering}: Mitigates \textbf{heterogeneity amplification} (driven by non-IID data, leading to \textbf{client drift}) by eliminating directional bias within the same version group, ensuring local updates across clients remain consistent in optimization direction.
	
	\item \textbf{Fairness-aware Aggregation}: Solves \textbf{unfair participation} (resulting from fast clients dominating aggregation, causing \textbf{unfair aggregation}) by adaptively balancing contributions from fast and slow clients, avoiding global model bias toward high-frequency participants.
	
\end{itemize}

The training pipeline of AlignFed consists of four key stages, which are detailed as follows:

\begin{enumerate}
	\item \textbf{Client-side Fine-tuning (\ding{172} in Fig.~\ref{framework}, Algorithm~\ref{alg:AlignFed}, Lines 3–7) :}
	
	Client $i$ downloads the global model with version $v_i$ and performs lightweight local fine-tuning via LoRA — a classic method for LLM federated fine-tuning due to its efficiency. Specifically, we freeze the pre-trained LLM backbone $W_{\text{backbone}}$ and only optimize two low-rank adapter matrices $A_i \in \mathbb{R}^{d \times r}$ and $B_i \in \mathbb{R}^{r \times d}$ (where $d$ is the LLM's hidden dimension, $r \ll d$ is the rank of LoRA). This design reduces the client's local parameter volume by over 95\% compared to full-model fine-tuning, making it feasible for resource-constrained edge devices~\cite{hu2022lora}.
	
	To avoid local feature drift in LLM's high-dimensional semantic space, we introduce a representation consistency regularization term:
	\begin{equation}
		\mathcal{L}_{\text{local}} = \mathcal{L}_{\text{task}} + \lambda \|\phi_i - \phi_g^{(v_i)}\|_2^2,
		\label{eq:local_loss}
	\end{equation}
	where $\mathcal{L}_{\text{task}}$ is the downstream task loss (e.g., cross-entropy for classification, negative log-likelihood for text generation), $\phi_i$ and $\phi_g^{(v_i)}$ are the feature representations extracted from the penultimate layer of the local/global LLM, and $\lambda \in [0.1, 1.0]$ balances task performance and global alignment. Unlike conventional regularization that constrains parameters directly, this term targets LLM's semantic representation — a more effective way to mitigate drift in large-scale models. As parameter-level constraints often fail to capture high-dimensional semantic consistency \cite{na2026semantic}, while semantic-level alignment can better preserve the global consistency of model cognition, especially in federated scenarios with non-IID data \cite{xiong2026step}.
	
	After completing local epochs (adapted to dataset size), the client computes the LoRA parameter increment $\Delta_i = (A_i - A_g^{(v_i)}, B_i - B_g^{(v_i)})$ (where $A_g^{(v_i)}, B_g^{(v_i)}$ are the global LoRA parameters at version $v_i$) and uploads only $\Delta_i$ and the version tag $v_i$ to the server. This lightweight upload minimizes communication overhead, a key requirement for edge-centric federated systems.
	
	\item \textbf{Version-aware Grouping (\ding{173} in Fig.~\ref{framework}, Algorithm~\ref{alg:AlignFed}, Lines 8–10) :}
	Upon receiving asynchronous updates $\{(\Delta_i, v_i)\}$, the server first groups them by their originating version to address staleness accumulation:
	\begin{equation}
		\mathcal{G}_k = \{ \Delta_i \mid v_i = k \}.
	\end{equation}
	This isolates updates trained on the same global version, avoiding mixed semantic spaces caused by direct aggregation of stale and fresh updates (e.g., updates from version $k=t-3$ mixed with version $k=t$).
	
	The server maintains a buffer for pending updates, with a heterogeneity-aware triggering rule for aggregation: trigger aggregation if either (1) the number of pending updates exceeds $M$ (balancing update diversity and timeliness), or (2) the elapsed time since last aggregation exceeds $\Delta T$ (preventing fast clients from waiting indefinitely). This rule adapts to variable upload speeds—e.g., if most clients are fast, aggregation triggers via condition (1); if many are slow, condition (2) ensures timely global model updates.
	
	\item \textbf{Intra-group Centering (\ding{174} in Fig.~\ref{framework}, Algorithm~\ref{alg:AlignFed}, Lines 11–12) :}
	This stage sequentially addresses heterogeneity amplification:
	
	Intra-group Centering eliminates directional bias caused by non-IID data within each version group (e.g., a group of industrial clients all pushing updates toward equipment-maintenance semantics), which mitigates client drift from heterogeneous data:
	\begin{equation}
		\tilde{\Delta}_i = \Delta_i - \frac{1}{|\mathcal{G}_k|}\sum_{\Delta_j \in \mathcal{G}_k} \Delta_j.
		\label{eq:centering}
	\end{equation}
	The centered increment $\tilde{\Delta}_i$ retains client-specific task knowledge while removing group-level shared bias, effectively suppressing \textbf{heterogeneity amplification} in asynchronous environments.
	
	\item \textbf{Cross-version Semantic Alignment (\ding{175} in Fig.~\ref{framework}, Algorithm~\ref{alg:AlignFed}, Lines 13–14) :}
	Building on the version-aware grouping (Stage 2), this step projects stale updates (from $\mathcal{G}_k$, $k < t$) into the current semantic space of $W_g^{(t)}$ to solve the "version mismatch" problem induced by staleness accumulation through Cross-version Semantic Alignment. Specifically, this step projects stale updates (from $\mathcal{G}_k$, $k < t$) into the current semantic space of $W_g^{(t)}$ to solve the "version mismatch" problem. We use a small server-side calibration set $\mathcal{D}_c$, which is constructed by randomly sampling 10\% of the public training dataset. Critically, $\mathcal{D}_c$ is completely separated from the test dataset used for final model evaluation — the test dataset is reserved exclusively for performance assessment and does not participate in any alignment or training processes.
	
	We learn a lightweight linear transformation $\mathcal{T}_{k \rightarrow t}: \mathbb{R}^{d \times r} \times \mathbb{R}^{r \times d} \rightarrow \mathbb{R}^{d \times r} \times \mathbb{R}^{r \times d}$ (applied to LoRA matrices $A$ and $B$ separately) to minimize the semantic discrepancy between versions:
	\begin{equation}
		\min_{\mathcal{T}_{k \rightarrow t}} \| f(W_g^{(t)}, \mathcal{D}_c) - f(W_g^{(k)} + \mathcal{T}_{k \rightarrow t}(\tilde{\Delta}_i), \mathcal{D}_c) \|_2^2,
		\label{eq:alignment}
	\end{equation}
	where $f(\cdot, \mathcal{D}_c)$ extracts the mean feature representation of $\mathcal{D}_c$ from the LLM's penultimate layer. The choice of linear transformation is deliberate for LLM scenarios: 
	(1) Computational efficiency: linear transformations have $\mathcal{O}(d \cdot r \cdot |\mathcal{D}_c|)$ complexity, which is negligible even for large-scale LLMs; 
	(2) Semantic preservation: LLM's feature space is approximately linear in the local update range, which is formally proved in Section \ref{sec:theoretical_analysis} (Theoretical Analysis); 
	(3) No extra client burden: unlike non-linear alignment methods (e.g., neural networks), linear transformations require no additional client communication or computation. 
	
	For the current version group $\mathcal{G}_t$ ($k=t$), $\mathcal{T}_{t \rightarrow t}$ is the identity transformation (no alignment needed).
	
	\item \textbf{Adaptive Fairness-aware Aggregation (\ding{176} in Fig.~\ref{framework}, Algorithm~\ref{alg:AlignFed}, Lines 15–16):}
	Computational heterogeneity leads to unfair participation: fast clients contribute significantly more updates than slow edge devices (e.g., IoT devices), dominating the global model. AlignFed addresses this via adaptive weights $\alpha_i$ that jointly account for three factors (all normalized to avoid scale bias):
	
	\begin{itemize}
		\item \textit{Freshness weight $\omega(v_i)$}: A decaying exponential function of staleness $\tau_i = t - v_i$ (time lag between client's version and current global version), defined as $\omega(v_i) = e^{-\gamma \cdot \tau_i}$ with $\gamma \in [0.1, 0.3]$. This down-weights highly stale updates (e.g., $\tau_i > 5$) that are semantically inconsistent with the current global model, while retaining valid updates from slightly delayed clients.
		
		\item \textit{Update strength weight}: Defined as $1/(\|\tilde{\Delta}_i\|_2 + \epsilon)$ (where $\epsilon = 10^{-8}$ to avoid division by zero), this term down-weights outlier updates with large deviation from the group mean (noisy or drifting clients)and emphasizes stable, consistent updates that align with the group pattern.

		\item \textit{Fairness weight $\xi_i$}: Inversely proportional to client $i$'s historical participation frequency to prevent over-representation. Specifically, $\text{num\_uploads}_i$ is the number of updates that client $i$ has uploaded to the server before the current aggregation round, and $\xi_i = 1/\sqrt{\text{num\_uploads}_i + 1}$. This design ensures that fast clients are not excluded, yet their dominance is curtailed, thereby striking a balance between fairness and update quality.
	\end{itemize}
	
	The normalized aggregation weight $\alpha_i$ (ensuring sum of weights = 1) is:
	\begin{equation}
		\label{eq:weight_def}
		\alpha_i = \frac{\omega(v_i) \cdot \xi_i}{\sum_j \omega(v_j) \cdot \xi_j / (\|\tilde{\Delta}_j\|_2 + \epsilon)} \cdot \frac{1}{\|\tilde{\Delta}_i\|_2 + \epsilon},
	\end{equation}
	and the global model update rule is:
	\begin{equation}
		W_g^{(t+1)} = W_g^{(t)} + \sum_i \alpha_i \, \mathcal{T}_{v_i \rightarrow t}(\tilde{\Delta}_i),
		\label{eq:aggregation}
	\end{equation}
	
	This weighting ensures that: (1) fast clients contribute fresh, strong updates but are not overrepresented; (2) slow clients’ valid updates are still considered (if fresh enough); (3) small-dataset clients are not marginalized — effectively solving the \textbf{unfair participation} problem.

\end{enumerate}

\begin{algorithm}[t]
	\caption{AlignFed Framework}
	\label{alg:AlignFed}
	\begin{algorithmic}[1]
		\STATE Initialize global model $W_g^{(0)}$
		\FOR{each round $t=1,2,\dots$}
		\FOR{each client $i$ in parallel}
		\STATE Pull global model $W_g^{(v_i)}$
		\STATE Perform local fine-tuning using Eq.~\eqref{eq:local_loss}
		\STATE Upload $(\Delta_i, v_i)$
		\ENDFOR
		\IF{ $|\mathcal{U}_{\text{pending}}| \ge M$ \textbf{or} $\Delta t > \Delta T$ }
		\STATE Group updates by version: $\mathcal{G}_k$
		\ENDIF
		\FOR{each group $\mathcal{G}_k$}
		\STATE Center updates using Eq.~\eqref{eq:centering}
		\STATE Align to current version using Eq.~\eqref{eq:alignment}
		\ENDFOR
		\STATE Aggregate aligned updates using Eq.~\eqref{eq:aggregation}
		\STATE Broadcast new model $W_g^{(t+1)}$
		\ENDFOR
	\end{algorithmic}
\end{algorithm}	
\section{Theoretical Analysis and Justification}
\label{sec:theoretical_analysis}
The following analysis theoretically justifies the three key components of AlignFed: version-aware grouping (staleness control), semantic alignment (drift reduction), and fairness-aware aggregation (bias mitigation).
\subsection{Complexity and Discussion}

Unlike conventional asynchronous federated learning methods (e.g., FedAsync, FedBuff) that transmit full model parameters in each round, AlignFed operates in a lightweight federated fine-tuning paradigm. Each client only uploads low-rank parameter increments, typically accounting for less than 1--5\% of the full model size \cite{zhang2025lori}, which substantially reduces communication overhead.

The additional computational cost mainly arises from the cross-version semantic alignment 
step on the server, where lightweight linear transformations are estimated using a small 
calibration set $\mathcal{D}_c$. This process scales linearly with the number of model 
layers $D_{\text{layer}}$ and calibration samples $|\mathcal{D}_c|$, i.e.,
\begin{equation}
	\mathcal{O}(D_{\text{layer}} \cdot |\mathcal{D}_c|).
\end{equation}
Since $|\mathcal{D}_c|$ is typically small and $L$ is independent of client scale, the overhead remains negligible even for large-scale edge environments. Compared to non-linear alignment methods (e.g., neural network-based mapping), AlignFed significantly reduces server-side computation while maintaining comparable alignment quality, making it more practical for resource-constrained distributed systems.

Furthermore, the intra-group centering step adds minimal computational cost ($\mathcal{O}(|\mathcal{G}_k|)$ per version group $\mathcal{G}_k$) and requires no additional communication, as it is performed locally on the server using existing client updates. The fairness-aware aggregation step also operates in $\mathcal{O}(M)$ time (where $M$ is the number of pending updates), ensuring the overall framework scales efficiently with the number of clients.

\subsection{Theoretical Analysis}
We provide theoretical guarantees for AlignFed, focusing on how it mitigates the three core challenges (model drift, client drift, and unfair aggregation) in asynchronous federated learning. Let $W_g^{(t)}$ denote the global model at version $t$, client $i$ fine-tunes from a stale version $W_g^{(v_i)}$ ($v_i \le t$) with staleness $\tau_i = t - v_i$, and the server aggregates aligned updates via Eq.~\eqref{eq:aggregation}.

\textbf{Assumptions.}
We adopt standard assumptions for asynchronous FL \cite{forootani2025asynchronous} and introduce additional assumptions tailored to AlignFed:
\begin{itemize}
	\item \textbf{(A1) $L$-smoothness:} For each client $i$, the local objective function $F_i(W)$ is $L$-smooth, i.e., $\|\nabla F_i(W_1) - \nabla F_i(W_2)\| \le L\|W_1 - W_2\|$ for all model parameters $W_1, W_2$.
	\item \textbf{(A2) Bounded gradient variance:} The stochastic gradient of the local objective satisfies $\mathbb{E}\|g_i(W) - \nabla F_i(W)\|^2 \le \sigma^2$, where $g_i(W)$ is the stochastic gradient computed from local data.
	\item \textbf{(A3) Bounded staleness:} The staleness of all client updates is bounded by $\tau_{\max}$, i.e., $\tau_i = t - v_i \le \tau_{\max}$, ensured by the server's heterogeneity-aware triggering rule.
	\item \textbf{(A4) Contractive alignment:} The cross-version linear transformation $\mathcal{T}_{v_i \rightarrow t}$ is $\rho$-contractive ($\rho < 1$), i.e., $\|\mathcal{T}_{v_i \rightarrow t}(x) - \mathcal{T}_{v_i \rightarrow t}(y)\| \le \rho\|x - y\|$ for any two updates $x, y$.
	\item \textbf{(A5) Bounded update magnitude:} The LoRA parameter increments of all clients are bounded, i.e., $\|\Delta_i\| \le \Delta_{\max}$, which holds for fine-tuning with weight decay and gradient clipping.
	\item \textbf{(A6) Fairness weight boundedness:} The fairness weight $\xi_i$ is bounded in $[0, 1]$, i.e., $0 < \xi_{\min} \le \xi_i \le 1$ for all $i$, ensuring no client is completely excluded from aggregation.
	\item \textbf{(A7) Linear feature space:} The model's feature space (extracted from the penultimate layer, denoted $f(\cdot, \mathcal{D}_c)$ for calibration set $\mathcal{D}_c$) is approximately linear within the local update range—a property theoretically and empirically validated for LLMs under low-rank fine-tuning (e.g., LoRA) \cite{jiang2024origins}. Formally, for any two model parameters $W_1 = W_g^{(k)} + \tilde{\Delta}_i$ and $W_2 = W_g^{(k)} + \tilde{\Delta}_j$ (within the same version group $\mathcal{G}_k$), the feature mapping satisfies:
	\begin{equation}
		\|f(W_1, \mathcal{D}_c) - f(W_2, \mathcal{D}_c) - \mathcal{T}_{k \rightarrow t}(W_1 - W_2)\| \le \epsilon_{\text{lin}},
	\end{equation}
	where $\epsilon_{\text{lin}}$ is a small constant (negligible for local LoRA updates), and $\mathcal{T}_{k \rightarrow t}$ is the linear transformation used for cross-version alignment.
\end{itemize}

\begin{lemma}[Bounded Asynchronous Drift (Mitigates Staleness Accumulation)]
	\label{lemma:drift}
	Under Assumptions (A1)–(A3) and (A5), the model deviation induced by staleness is bounded by:
	\begin{equation}
	\|W_g^{(t)} - W_g^{(v_i)}\| \le \eta \tau_{\max} \Delta_{\max}.
\end{equation}
	where $\eta$ is the global learning rate. After applying cross-version semantic alignment (Assumptions A4 and A7), the aligned update error is further bounded by:
	\begin{equation}
	\|\mathcal{T}_{v_i \rightarrow t}(\tilde{\Delta}_i) - \Delta_i^*\|
	\le \rho \eta L \tau_{\max} \Delta_{\max} + \epsilon_{\text{lin}}.
\end{equation}
	where $\Delta_i^*$ is the ideal update aligned to the current version $t$, and $\tilde{\Delta}_i$ is the centered increment from Eq.~\eqref{eq:centering}.

\textit{Proof Sketch:}
The first bound characterizes the global model deviation caused by asynchronous staleness, which is accumulated by at most $\tau_{\max}$ bounded update steps.
The second bound further illustrates that the stale update can be corrected toward the ideal fresh update via the $\rho$-contractive semantic alignment, with a small constant error $\epsilon_{\text{lin}}$ introduced by linear feature approximation.
This lemma theoretically guarantees that AlignFed effectively suppresses staleness-induced drift and maintains accurate semantic information during cross-version adaptation.
	\textit{Detailed proof is provided in Appendix~\ref{app:proof_1}.}
	
\end{lemma}
\begin{lemma}[Bounded Client Drift (Mitigates Heterogeneity Amplification)]
	\label{lemma:client_drift}
	Under Assumptions (A1) and (A5), the intra-group centering (Eq.~\eqref{eq:centering}) and representation regularization (Eq.~\eqref{eq:local_loss}) ensure:
	\begin{equation}
		\|\tilde{\Delta}_i\| \le \min\left\{\Delta_{\max},\; \sqrt{\frac{\mathcal{L}_{\text{local}} - \mathcal{L}_{\text{task}}}{\lambda}}\right\},
	\end{equation}
	where $\tilde{\Delta}_i$ is the centered local update within version group $\mathcal{G}_k$.
	Furthermore, the gradient deviation induced by local update is bounded via $L$-smoothness as:
	\begin{equation}
		\|\nabla F_i(W_g^{(v_i)} + \tilde{\Delta}_i) - \nabla F_i(W_g^{(v_i)})\|
		\le L \|\tilde{\Delta}_i\|.
	\end{equation}

\textit{Proof Sketch:}
Intra-group centering eliminates the group-level common bias, yielding $\mathbb{E}[\tilde{\Delta}_i \mid \mathcal{G}_k] = 0$.
The representation regularization term explicitly constrains the norm of the centered update $\tilde{\Delta}_i$, which is further bounded by $\Delta_{\max}$ from Assumption (A5).
By $L$-smoothness (A1), the gradient deviation caused by local fine-tuning is linearly bounded by the update norm.
This lemma ensures that AlignFed effectively suppresses client drift and mitigates heterogeneity amplification across non-IID clients.
	\textit{Detailed proof is provided in Appendix~\ref{app:proof_2}.}
\end{lemma}

\begin{lemma}[Fair Contribution Bounding (Mitigates Unfair Participation)]
	\label{lemma:fairness}
	Under Assumptions (A5) and (A6), the fairness-aware aggregation weight satisfies:
		\begin{equation}
		\alpha_i = \frac{\omega_i \xi_i / (\|\tilde{\Delta}_i\|_2+\epsilon)}
		{\sum_j \omega_j \xi_j / (\|\tilde{\Delta}_j\|_2+\epsilon)}.
	\end{equation}
	The lower bound is given by
	\begin{equation}
		\alpha_i \ge \frac{\omega_{\min}\xi_{\min}}{\sum_j \omega_j \xi_j / (\|\tilde{\Delta}_j\|_2+\epsilon)} \cdot \frac{1}{\Delta_{\max}+\epsilon},
	\end{equation}
	where $\omega_{\min}=e^{-\gamma\tau_{\max}}$ is the minimum freshness weight, $\xi_{\min}$ is the lower bound of fairness weight, and $\epsilon>0$ is a small constant for numerical stability.
	Furthermore, the weight ratio between any two clients is bounded by:
	\begin{equation}
		\frac{\alpha_i}{\alpha_j} \le \frac{\omega_{\max}\xi_{\max}}{\omega_{\min}\xi_{\min}} \cdot \frac{\|\tilde{\Delta}_j\|_2+\epsilon}{\|\tilde{\Delta}_i\|_2+\epsilon},
	\end{equation}
	where $\omega_{\max}=1$ denotes the weight for fresh updates.

\textit{Proof Sketch:}
The lower bound of $\alpha_i$ ensures that slow or infrequent clients always maintain a non-negligible contribution in aggregation.
The bounded weight ratio restricts the dominance of fast or frequent clients, preventing weight imbalance across clients.
By jointly constraining freshness and fairness scores, AlignFed achieves stable and balanced aggregation under asynchronous settings.
	\textit{Detailed proof is provided in Appendix~\ref{app:proof_3}.}
\end{lemma}
	
\begin{theorem}[Ergodic Convergence (Overall Theoretical Guarantee)]
	\label{theorem:convergence}
	Under Assumptions (A1)--(A7) and learning rate $\eta \le 1/(2L)$, the ergodic gradient norm over $T$ iterations satisfies:
	\begin{equation}
		\begin{aligned}
			\frac{1}{T}\sum_{t=0}^{T-1}\mathbb{E}\|\nabla F(W_g^{(t)})\|^2
			\le\;& \mathcal{O}\!\left(\frac{1}{T}\right)
			+ \mathcal{O}\!\left(\eta^2 L^2 \tau_{\max}^2 \rho^2 \Delta_{\max}^2\right) \\
			&+ \mathcal{O}\!\left(\frac{\sigma^2}{N}\right)
			+ \mathcal{O}(\epsilon_{\text{lin}}^2),
		\end{aligned}
	\end{equation}
	where $N$ is the total number of clients, $\sigma^2$ is the stochastic gradient variance, and $\epsilon_{\text{lin}}$ denotes the residual error of linear semantic alignment.
	The four terms respectively represent convergence optimality, asynchronous staleness error, client sampling variance, and cross-version alignment error.

\textit{Proof Sketch:}
By applying the $L$-smoothness descent lemma and Young's inequality, we establish the basic convergence recursion.
The bounded staleness from Lemma~\ref{lemma:drift}, bounded client drift from Lemma~\ref{lemma:client_drift}, and fair aggregation from Lemma~\ref{lemma:fairness} together ensure that all error terms are well constrained.
The final bound achieves standard $\mathcal{O}(1/T)$ convergence rate under asynchronous federated settings, which theoretically validates the effectiveness and stability of AlignFed.
	\textit{Detailed proof is provided in Appendix~\ref{app:proof_4}.}
\end{theorem}

The theoretical results highlight three key insights:
1. Staleness mitigation: The contractive alignment ($\rho < 1$) reduces the staleness penalty by a factor of $\rho^2$, outperforming conventional methods that only penalize stale updates via weight decay. Additionally, the approximate linearity of the feature space (A7) ensures the linear transformation preserves feature information (i.e., semantic preservation), confirming the design choice in the framework section—linear transformations are sufficient to align stale updates without losing critical task-related information.
2. Heterogeneity resilience: Intra-group centering and representation regularization bound client drift, ensuring the global model does not diverge due to non-IID data.
3. Fairness guarantee: The bounded weight ratio prevents fast clients from dominating aggregation, while the lower bound of $\alpha_i$ ensures slow clients' valid updates are retained.

Notably, AlignFed achieves the same asymptotic convergence rate as synchronous federated learning and outperforms naive asynchronous methods (e.g., FedAsync) by reducing the staleness penalty via contractive alignment. This confirms that the proposed multi-stage alignment mechanism effectively addresses the three core challenges without sacrificing convergence speed or scalability.

	\section{Experiments}

We evaluate AlignFed across diverse federated edge scenarios to comprehensively validate its effectiveness, stability, and generalization.
Specifically, we conduct four sets of experiments:
(1) convergence analysis to compare synchronous and asynchronous training;
(2) robustness testing under varying staleness tolerance levels;
(3) performance comparison with state-of-the-art federated LLM fine-tuning methods;
and (4) cross-model validation on a second LLM backbone to verify model-agnostic behavior.

\subsection{Experimental Setup}

All experiments were conducted on a single NVIDIA A100 GPU using two large language models: Llama3-8B and Qwen3-8B. 
We follow the FederatedScope-LLM framework~\cite{kuang2024federatedscope} for system simulation, adopting its heterogeneity settings.
All models were trained in FP16 with a maximum sequence length of 650 and batch size of 1, emulating constrained edge devices. 
Client-to-server communication per round was limited to $\le 50$MB.

We evaluate on three datasets: GSM8K~\cite{cobbe2021training} (mathematical reasoning), 
CodeAlpaca~\cite{chaudhary2023code} (code generation), 
and Dolly~\cite{conover2023free} (instruction following).
Following prior work, each dataset was partitioned into 50 clients using a Dirichlet distribution with $\alpha=0.5$ to simulate non-IID federated environments.

For parameter-efficient fine-tuning, we apply LoRA updates on the query and value projection matrices. Each client performs local LoRA fine-tuning and uploads its lightweight update asynchronously to the server.

\subsection{Convergence Analysis on Llama3-8B}

Due to the heterogeneous computation speeds and non-IID data distributions across edge clients, synchronous FFT often suffers from delayed global updates and amplified client drift. 
To evaluate the impact of asynchrony under such conditions, we compare synchronous and asynchronous federated LoRA fine-tuning on Llama3-8B across three datasets.

\begin{figure}[!htbp]
	\centerline{\includegraphics[width=0.4\textwidth]{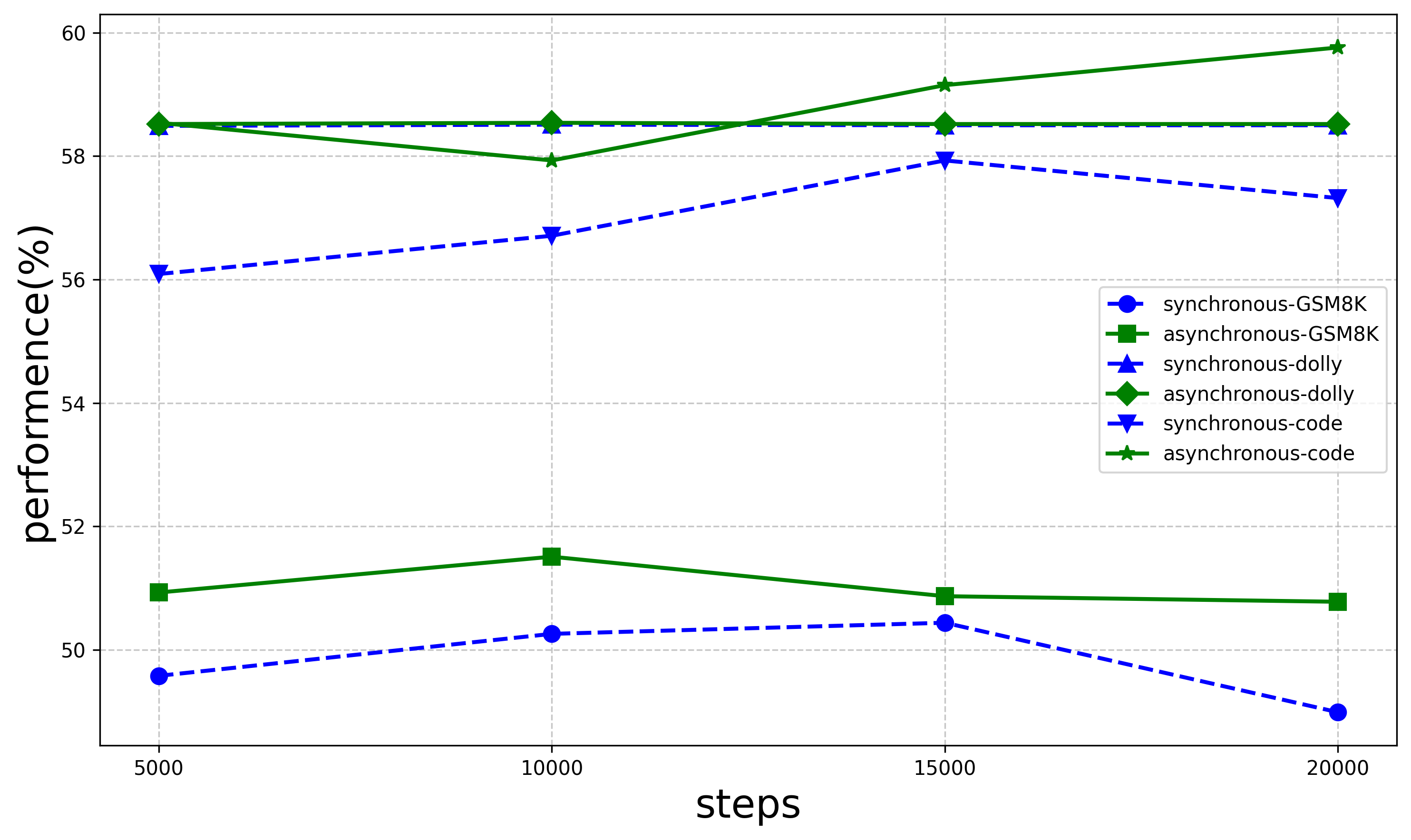}}
	\caption{Comparison of synchronous vs.\ asynchronous federated LoRA fine-tuning on Llama3-8B.}
	\label{shiyan}
\end{figure}

Fig.~\ref{shiyan} compares synchronous and asynchronous federated LoRA fine-tuning on Llama3-8B across the three datasets. 
Because asynchronous aggregation updates the global model whenever a subset of clients (about 30 per round) returns updates—rather than waiting for all 50 clients as in synchronous FFT—it achieves significantly faster and more stable convergence.

On \textbf{GSM8K}, synchronous training shows strong oscillations (49.58→50.44→48.99) caused by model version drift, 
while asynchronous updates maintain both higher and more stable performance (50.93–51.51).
On \textbf{Dolly}, both methods converge to similar accuracy ($\approx 58.5\%$), consistent with the lower gradient variance of instruction-following tasks.

The largest gain appears on \textbf{CodeAlpaca}: asynchronous fine-tuning steadily improves to 59.76, while synchronous optimization remains noisy and suboptimal. 
This confirms that code-generation tasks, which exhibit stronger client-specific specialization, are particularly vulnerable to gradient inconsistency under synchronous FedAvg~\cite{mcmahan2017communication}.

Overall, the results demonstrate that asynchronous federated fine-tuning reduces stale gradient effects and achieves faster, smoother convergence under real-world heterogeneity.

\subsection{Robustness to Staleness}

We further evaluate the robustness of AlignFed by varying the \emph{maximum staleness tolerance} parameter $\tau_{\max}$, 
which restricts the maximum allowed version gap ($\tau_i = t - v_i$) between a client's local model (version $v_i$) and the current global model (version $t$).

Table~\ref{tab:staleness} reports pass@1 and pass@10 on CodeAlpaca under $\tau_{\max} \in \{1,2,3,4,5\}$. 
AlignFed remains stable across all staleness levels. Performance fluctuations are small even at $\tau_{\max}=5$. This validates that the proposed multi-stage alignment mechanism effectively mitigates model drift caused by stale updates.

\begin{table}[htbp]
	\centering
	\caption{Robustness of AlignFed to staleness tolerance $\tau_{\max}$ on CodeAlpaca.}
	\label{tab:staleness}
	\begin{tabular}{c|cc}
		\hline
		\textbf{Staleness $\tau_{\max}$} & \textbf{pass@1} & \textbf{pass@10} \\
		\hline
		1 & 34.88 & 58.54 \\
		2 & 34.21 & 58.54 \\
		3 & 35.06 & 59.76 \\
		4 & 34.02 & 57.32 \\
		5 & 36.16 & 55.49 \\
		\hline
	\end{tabular}
\end{table}
\subsection{Comparison with State-of-the-Art Federated LLM Methods}
We further compare AlignFed with representative federated LLM fine-tuning baselines, including:
\begin{itemize}
	\item \textbf{FedAvg}~\cite{mcmahan2017communication}: Standard synchronous federated averaging baseline.
	\item \textbf{FedAvg+FedBuff}: FedAvg combined with FedBuff~\cite{nguyen2022fedbuff} to handle asynchronous updates.
	\item \textbf{FFA-LoRA}~\cite{sun2024improving}: State-of-the-art federated LoRA fine-tuning method focusing on client drift mitigation.
	\item \textbf{FFA-LoRA+FedBuff}: FFA-LoRA integrated with asynchronous buffering mechanism.
	\item \textbf{FedSA-LoRA}~\cite{guo2025selective}: Federated semantic alignment method for LLM fine-tuning.
	\item \textbf{FedSA-LoRA+FedBuff}: FedSA-LoRA with asynchronous buffering.
\end{itemize}

To ensure fair comparison with existing federated LLM methods, we strictly follow the client partition setup \textbf{from FederatedScope-LLM~\cite{kuang2024federatedscope}}: GSM8K (3 IID clients), CodeAlpaca (9 non-IID clients by programming languages), and Dolly (8 task-specialized clients).
In contrast, we use \textbf{50 clients} in our convergence, robustness, and cross-model experiments to \textbf{simulate large-scale heterogeneous edge environments with numerous devices}, which better reflects real-world asynchronous federated learning scenarios and validates the scalability of AlignFed.

\begin{table}[htbp]
	\centering
	\caption{Performance comparison of AlignFed with state-of-the-art federated LLM methods on Llama3-8B.}
	\label{tab:main_results}
	\begin{tabular}{l|rrrr}
		\toprule
		\textbf{Method} & \textbf{GSM8K} & \multicolumn{2}{c}{\textbf{CodeAlpaca}} & \textbf{Dolly} \\
		& \textbf{Accuracy} & \textbf{pass@1} & \textbf{pass@10} & \textbf{Score} \\
		\midrule
		FedAvg                        & 50.44 &\textbf{ 37.93} & \textbf{59.76} & 58.49 \\
		FedAvg+FedBuff                & 50.42 & 37.50 & \textbf{59.76} & 57.89 \\
		FedAvg+AlignFed               & \textbf{50.87} & 37.13 & 59.15 & \textbf{58.52} \\
		\midrule
		FFA-LoRA                      & 50.72 & 39.57 & 60.37 & 58.21 \\
		FFA-LoRA+FedBuff              & 52.08 & 39.88 & 61.59 & 57.94 \\
		FFA-LoRA+AlignFed             & \textbf{52.69} & \textbf{40.30} & \textbf{62.80} & \textbf{58.32} \\
		\midrule
		FedSA-LoRA                    & 51.10 & 38.60 & 60.98 & 58.00 \\
		FedSA-LoRA+FedBuff            & 51.48 & 39.21 & 59.15 & 57.71 \\
		FedSA-LoRA+AlignFed           & \textbf{51.97} & \textbf{40.73} & \textbf{61.59} & \textbf{58.13} \\
		\bottomrule
	\end{tabular}
\end{table}

AlignFed consistently improves performance across all methods and datasets.  Among all variants, FFA-LoRA+AlignFed achieves the best overall performance with 52.69\% on GSM8K, 40.30\% pass@1 and 62.80\% pass@10 on CodeAlpaca, outperforming all other baselines. On CodeAlpaca, FedSA-LoRA+AlignFed significantly improves pass@1 to 40.73\% and pass@10 to 61.59\%, surpassing both FedSA-LoRA and FedSA-LoRA+FedBuff, which demonstrates that AlignFed effectively mitigates model and client drift caused by asynchronous model updates, thereby improving generation stability.

On Dolly, all AlignFed-enhanced methods maintain or slightly improve performance without obvious degradation, showing that multi-stage alignment preserves the generalization ability for instruction-tuning tasks. Additionally, AlignFed consistently outperforms FedBuff-based variants in most metrics, indicating that targeted multi-stage alignment is more effective than simple buffering for mitigating staleness and heterogeneity in asynchronous federated LLM fine-tuning.

\subsection{Cross-Model Generalization: Qwen3-8B}

To verify that AlignFed is model-agnostic, we perform the same experiments on Qwen3-8B. 
Table~\ref{tab:qwen} summarizes the final synchronous vs.\ asynchronous performance.

Across all datasets, the asynchronous variant slightly but consistently matches or outperforms synchronous method. 
The improvements are most evident on CodeAlpaca, where pass@1 and pass@10 both increase. 
These results confirm that AlignFed generalizes effectively to different LLM families, including multilingual architectures such as Qwen3-8B.

\begin{table}[h!]
	\centering
	\caption{Performance comparison on Qwen3-8B (synchronous vs.\ asynchronous).}
	\label{tab:qwen}
	\begin{tabular}{lcc}
		\toprule
		\textbf{Dataset / Metric} & \textbf{Sync} & \textbf{Async} \\
		\midrule
		GSM8K (Accuracy)         & 87.41 & \textbf{88.32 }\\
		\midrule
		CodeAlpaca (pass@1)      & 53.29 & \textbf{53.66 }\\
		CodeAlpaca (pass@10)     & 71.34 & \textbf{72.56} \\
		\midrule
		Dolly (MMLU score)       & 66.65 & \textbf{66.66} \\
		\bottomrule
	\end{tabular}
\end{table}

These findings further demonstrate that AlignFed is not tied to a specific backbone and provides stable improvements across heterogeneous LLM architectures.

\subsection{Summary of Experimental Findings}

Across all experiments, AlignFed consistently and reliably demonstrated enhanced performance in the asynchronous federated fine-tuning of Large Language Models (LLMs).
Convergence analysis confirmed that employing our proposed multi-stage alignment mechanism for asynchronous training facilitates a faster and more stable optimization process compared to synchronous schemes.
Robustness tests indicated that AlignFed maintains robust performance—without experiencing significant fluctuations—even when confronted with severe update staleness.
Comparisons with state-of-the-art methods further validated that AlignFed integrates effectively with existing advanced federated LLM fine-tuning techniques, while delivering significantly superior performance compared to other asynchronous baseline methods.
Cross-model experiments conducted using the Qwen3-8B model confirmed that AlignFed possesses universal generalization capabilities independent of specific model architectures.
Collectively, these results demonstrate that AlignFed effectively addresses the critical challenges facing asynchronous federated learning in real-world scenarios: model drift caused by the accumulation of staleness, client drift resulting from data heterogeneity, and unfair aggregation driven by the dominance of fast clients.

\section{Conclusion}

In this work, we introduced AlignFed, an asynchronous federated fine-tuning framework tailored for large language models in heterogeneous edge environments. 
Motivated by the limitations of synchronous FFT—particularly the straggler effect and inefficient resource utilization—and the inadequacy of existing asynchronous FL methods in handling LLM-specific challenges, AlignFed is designed to address three fundamental issues: staleness-induced model drift, heterogeneity-amplified client drift, and unfair aggregation due to imbalanced client participation.

To this end, AlignFed incorporates a multi-stage alignment mechanism that unifies: (i) version-aware update grouping to control staleness accumulation, (ii) cross-version semantic alignment to ensure representation consistency across model versions, and (iii) fairness-aware aggregation to balance client contributions under heterogeneous participation. 
By explicitly aligning updates in both temporal and semantic dimensions, AlignFed enables stable and consistent model evolution under asynchronous training.

Extensive experiments on multiple datasets and LLM backbones demonstrate that AlignFed achieves faster and more stable convergence, improved robustness under non-IID data distributions, and more balanced performance across heterogeneous clients compared with synchronous FFT baselines. 

Overall, this work highlights the critical role of joint temporal--semantic alignment in asynchronous federated LLM fine-tuning, and provides a practical pathway toward scalable, efficient, and fair model adaptation for real-world edge intelligence systems.

\bibliographystyle{IEEEtran}
\bibliography{reference}

\clearpage  

\appendix
\section{Appendix: Detailed Proofs}

\subsection{Proof of Lemma~\ref{lemma:drift}}
\label{app:proof_1}
\begin{proof}
	\textbf{Step 1.}
	From the global update rule:
	\begin{equation}
		W_g^{(k+1)} = W_g^{(k)} + \eta \Delta_g^{(k)}.
	\end{equation}
	By (A5) and convex combination:
	\begin{equation}
		\|\Delta_g^{(k)}\|
		\le \sum_i \alpha_i \|\tilde{\Delta}_i\|
		\le \Delta_{\max}.
	\end{equation}
	Thus,
	\begin{equation}
		\|W_g^{(t)}-W_g^{(v_i)}\|
		\le \sum_{k=v_i}^{t-1} \eta \Delta_{\max}
		= \eta \tau_i \Delta_{\max}
		\le \eta \tau_{\max}\Delta_{\max}.
	\end{equation}
	
	\textbf{Step 2.}
	By $L$-smoothness (A1):
	\begin{equation}
		\|\tilde{\Delta}_i - \Delta_i^*\|
		\le L \|W_g^{(t)} - W_g^{(v_i)}\|
		\le \eta L \tau_{\max} \Delta_{\max}.
	\end{equation}
	By triangle inequality:
	\begin{align}
		&\|\mathcal{T}_{v_i\rightarrow t}(\tilde{\Delta}_i)-\Delta_i^*\| \\
		\le\;&
		\|\mathcal{T}(\tilde{\Delta}_i)-\mathcal{T}(\Delta_i^*)\|
		+ \|\mathcal{T}(\Delta_i^*)-\Delta_i^*\|.
	\end{align}
	Using contractivity (A4) and linear error (A7):
	\begin{equation}
		\le \rho \|\tilde{\Delta}_i-\Delta_i^*\| + \epsilon_{\text{lin}}.
	\end{equation}
	Substituting the bound yields:
	\begin{equation}
		\le \rho \eta L \tau_{\max} \Delta_{\max} + \epsilon_{\text{lin}}.
	\end{equation}
\end{proof}

\subsection{Proof of Lemma~\ref{lemma:client_drift}}
\label{app:proof_2}
\begin{proof}
	From representation regularization:
	\begin{equation}
		\lambda \|\tilde{\Delta}_i\|^2 \le \mathcal{L}_{\text{local}} - \mathcal{L}_{\text{task}}.
	\end{equation}
	Rearranging gives:
	\begin{equation}
		\|\tilde{\Delta}_i\|
		\le \sqrt{\frac{\mathcal{L}_{\text{local}}-\mathcal{L}_{\text{task}}}{\lambda}}.
	\end{equation}
	Combining with (A5):
	\begin{equation}
		\|\tilde{\Delta}_i\| \le \min\left\{\Delta_{\max},\; \sqrt{\frac{\mathcal{L}_{\text{local}}-\mathcal{L}_{\text{task}}}{\lambda}}\right\}.
	\end{equation}
	By $L$-smoothness (A1):
	\begin{equation}
		\|\nabla F_i(W+\tilde{\Delta}_i)-\nabla F_i(W)\|
		\le L\|\tilde{\Delta}_i\|.
	\end{equation}
\end{proof}

\subsection{Proof of Lemma~\ref{lemma:fairness}}
\label{app:proof_3}
\begin{proof}
	\textbf{Step 1 (Lower bound).}
	By definition:
	\begin{equation}
		\alpha_i =
		\frac{\omega_i \xi_i / (\|\tilde{\Delta}_i\|_2+\epsilon)}
		{\sum_j \omega_j \xi_j / (\|\tilde{\Delta}_j\|_2+\epsilon)}.
	\end{equation}
	Using $\omega_i\ge\omega_{\min}$, $\xi_i\ge\xi_{\min}$, $\|\tilde{\Delta}_i\|_2\le\Delta_{\max}$:
	\begin{equation}
		\alpha_i \ge
		\frac{\omega_{\min}\xi_{\min}}{\Delta_{\max}+\epsilon}
		\Bigg/
		\sum_j \frac{\omega_j \xi_j}{\|\tilde{\Delta}_j\|_2+\epsilon}.
	\end{equation}
	
	\textbf{Step 2 (Ratio bound).}
	Direct substitution gives:
	\begin{align}
		\frac{\alpha_i}{\alpha_j}
		&=
		\frac{\omega_i \xi_i}{\omega_j \xi_j}
		\cdot
		\frac{\|\tilde{\Delta}_j\|_2+\epsilon}{\|\tilde{\Delta}_i\|_2+\epsilon} \\
		&\le
		\frac{\omega_{\max}\xi_{\max}}{\omega_{\min}\xi_{\min}}
		\cdot
		\frac{\|\tilde{\Delta}_j\|_2+\epsilon}{\|\tilde{\Delta}_i\|_2+\epsilon}.
	\end{align}
\end{proof}
\subsection{Proof of Theorem~\ref{theorem:convergence}}
\label{app:proof_4}
\begin{proof}
	The expectation $\mathbb{E}$ is taken over client sampling, stochastic gradient noise, and staleness.
	
	\textbf{Step 1 (Smoothness descent).} 
	By $L$-smoothness (A1), we have:
	\begin{equation}
		F(W^{t+1})
		\le F(W^t)
		- \eta \langle \nabla F(W^t), \Delta_g^{(t)} \rangle
		+ \frac{L\eta^2}{2}\|\Delta_g^{(t)}\|^2.
	\end{equation}
	
	\textbf{Step 2 (Inner product control).} 
	Applying Young's inequality:
	\begin{equation}
		-\langle a,b\rangle \le \frac{1}{2}\|a\|^2 + \frac{1}{2}\|b\|^2,
	\end{equation}
	we obtain:
	\begin{equation}
		\langle \nabla F(W^t), \Delta_g^{(t)} \rangle
		\ge -\frac{1}{2}\|\nabla F(W^t)\|^2 - \frac{1}{2}\|\Delta_g^{(t)}\|^2.
	\end{equation}
	
	Substituting into Step 1 and rearranging yields:
	\begin{equation}
		\|\nabla F(W^t)\|^2
		\le
		\frac{2(F(W^t)-F(W^{t+1}))}{\eta}
		+ (L\eta + 1)\|\Delta_g^{(t)}\|^2. 
	\end{equation}
	
	\textbf{Step 3 (Error decomposition).} 
	Decompose the aligned update:
	\begin{equation}
		\mathcal{T}_{v_i \to t}(\tilde{\Delta}_i)
		= \Delta_i^* + e_i^{\text{stale}} + e_i^{\text{align}},
	\end{equation}
	where:
	\begin{itemize}
		\item $\Delta_i^*$: ideal update at $W_g^{(t)}$, satisfying $\mathbb{E}[\Delta_i^*] = \nabla F(W_g^{(t)})$ and $\mathbb{E}\|\Delta_i^* - \nabla F(W_g^{(t)})\|^2 \le \sigma^2$ (A2); 
		\item $e_i^{\text{stale}}$: bounded staleness error from Lemma~\ref{lemma:drift};
		\item $e_i^{\text{align}}$: alignment residual with $\|e_i^{\text{align}}\| \le \epsilon_{\text{lin}}$ (A7).
	\end{itemize}
	
	The aggregated update satisfies
	\begin{equation}
		\Delta_g^{(t)} = \sum_i \alpha_i \Delta_i^* + \sum_i \alpha_i e_i^{\text{stale}} + \sum_i \alpha_i e_i^{\text{align}},
	\end{equation}
	where $\alpha_i \ge 0$ and $\sum_i \alpha_i = 1$ (Lemma~\ref{lemma:fairness}).
	
	Using the inequality $\|x+y+z\|^2 \le 3(\|x\|^2+\|y\|^2+\|z\|^2)$, we obtain:
	\begin{equation}
		\|\Delta_g^{(t)}\|^2
		\le 3\left(
		\left\|\sum_i \alpha_i \Delta_i^*\right\|^2
		+ \left\|\sum_i \alpha_i e_i^{\text{stale}}\right\|^2
		+ \left\|\sum_i \alpha_i e_i^{\text{align}}\right\|^2
		\right). 
	\end{equation}
	
	We bound each term separately:
	
	\textbf{(i) Variance term.}
	Since $\Delta_i^*$ are unbiased estimators and independent across clients,
	\begin{equation}
		\mathbb{E}\left\|\sum_i \alpha_i \Delta_i^*\right\|^2
		\le \|\nabla F(W_g^{(t)})\|^2 + \frac{\sigma^2}{N}, 
	\end{equation}
	where we use $\mathbb{E}[\alpha_i]=1/N$ and $\sum_i \alpha_i^2 \le 1/N$.
	
	\textbf{(ii) Staleness error.}
	From Lemma~\ref{lemma:drift},
	\begin{equation}
		\|e_i^{\text{stale}}\|^2
		\le \mathcal{O}(\eta^2 L^2 \tau_{\max}^2 \rho^2 \Delta_{\max}^2).
	\end{equation}
	Thus,
	\begin{equation}
		\left\|\sum_i \alpha_i e_i^{\text{stale}}\right\|^2
		\le \mathcal{O}(\eta^2 L^2 \tau_{\max}^2 \rho^2 \Delta_{\max}^2).
	\end{equation}
	
	\textbf{(iii) Alignment error.}
	\begin{equation}
		\left\|\sum_i \alpha_i e_i^{\text{align}}\right\|^2
		\le \epsilon_{\text{lin}}^2.
	\end{equation}
	
	Combining all bounds:
	\begin{equation}
			\begin{aligned}
		\mathbb{E}\|\Delta_g^{(t)}\|^2
		&\le
		\mathcal{O}\!\left(\|\nabla F(W_g^{(t)})\|^2\right)\\
		&+ \mathcal{O}\!\left(\frac{\sigma^2}{N}\right)\\
		&+ \mathcal{O}(\eta^2 L^2 \tau_{\max}^2 \rho^2 \Delta_{\max}^2)
		+ \mathcal{O}(\epsilon_{\text{lin}}^2). 
				\end{aligned}
	\end{equation}
	
	\textbf{Step 4 (Plug in and absorb).}
	
	Substitute into Step 2:
	\begin{equation}
		\begin{aligned}
			\mathbb{E}\|\nabla F(W^t)\|^2
			\le&
			\frac{2\mathbb{E}[F(W^t)-F(W^{t+1})]}{\eta}\\
			&+ (L\eta+1)\cdot \mathcal{O}(\|\nabla F(W^t)\|^2) \\
			&+ \mathcal{O}\!\left(\frac{\sigma^2}{N}\right)
			+ \mathcal{O}(\eta^2 L^2 \tau_{\max}^2 \rho^2 \Delta_{\max}^2)
			+ \mathcal{O}(\epsilon_{\text{lin}}^2).
		\end{aligned}
	\end{equation}
	
	Using $\eta \le \frac{1}{2L}$, we have $(L\eta+1)\cdot c < 1$ for some constant $c$, 
	thus the gradient term can be absorbed to the left-hand side. 
	
	\textbf{Step 5 (Telescoping sum).}
	
	Summing over $t=0,\dots,T-1$:
	\begin{equation}
		\sum_{t=0}^{T-1} \mathbb{E}[F(W^t)-F(W^{t+1})]
		\le F(W^0)-F^*,
	\end{equation}
	where $F^*$ is the lower bound of $F$.
	
	Dividing by $T$, we obtain:
	\begin{equation}
		\begin{aligned}
			\frac{1}{T}\sum_{t=0}^{T-1}\mathbb{E}\|\nabla F(W_g^{(t)})\|^2
			\le\;& \mathcal{O}\!\left(\frac{1}{T}\right)
			+ \mathcal{O}\!\left(\eta^2 L^2 \tau_{\max}^2 \rho^2 \Delta_{\max}^2\right) \\
			&+ \mathcal{O}\!\left(\frac{\sigma^2}{N}\right)
			+ \mathcal{O}(\epsilon_{\text{lin}}^2),
		\end{aligned}
	\end{equation}

\end{proof}

\end{document}